# Large Language Models are as persuasive as humans, but how? About the cognitive effort and moral-emotional language of LLM arguments.

**Carlos Carrasco-Farré**
Information Systems Department
Toulouse Business School

**Abstract.** Large Language Models (LLMs) are already as persuasive as humans. However, we know very little about *how* they do it. This paper investigates the persuasion strategies of LLMs, comparing them with human-generated arguments. Using a dataset of 1,251 participants in an experiment, we compare the persuasion strategies of LLM-generated and human-generated arguments through measures of cognitive effort (lexical and grammatical complexity) and moral-emotional language (sentiment and morality). Our results indicate that LLMs produce arguments that require higher cognitive effort, exhibiting more complex grammatical and lexical structures than human counterparts. Additionally, LLMs demonstrate a significant propensity to engage more deeply with moral language, utilizing both positive and negative moral foundations more frequently than humans. In contrast with previous research, no significant difference was found in the emotional content produced by LLMs and humans. The fact that we show that there is no equivalence in process despite equivalence in outcome, contributes to the emergent knowledge regarding AI and persuasion, highlighting the dual potential of LLMs to both enhance and undermine informational integrity through persuasion strategies.

# 1. INTRODUCTION

Large Language Models (LLMs) can create content that is highly persuasive, equaling, or even surpassing (Hackenburga et al., 2023), the effectiveness of humans in convincing users about contentious political issues (Bai et al., 2023). These findings are raising concerns, since LLMs are capable of producing content as persuasive as original propaganda crafted by humans (Goldstein et al., 2024). Also, the persuasiveness of deceptive content is increased when LLMs can access personal information for tailoring the messages to specific audiences, allowing for cheap automation of persuasive misinformation at a huge scale, increasing not just its efficiency, but also its effectiveness (Costello et al., 2024; Matz et al., 2024). If that was not enough, these results are even more worrying when we take into account that the persuasion capabilities of LLMs is increasing as these models evolve (Durmus et al., 2024), with the potential of creating a perfect storm of misinformation (Galaz et al., 2023). However, despite the empirical evidence that LLMs are already as persuasive as humans, and likely to surpass it in next developments, we know very little about *how*.

To shed light into the research gap of how LLMs are as persuasive as humans, we rely on previous evidence about communication strategies for persuasiveness showing that cognitive effort, and moral-emotional language is associated with higher persuasiveness. For example, existing evidence indicates that each additional negative word in a headline boost click-through rate by 2.3% (Robertson et al., 2023). Moreover, previous evidence shows that reduced cognitive effort to process content is associated with viral misinformation (Carrasco-Farré, 2022). Lastly, the persuasiveness of "moral-emotional" language suggests that high-arousal, morally charged, and emotional rhetoric is highly persuasive (Brady et al., 2017; Rathje et al., 2021).

In order to test whether LLMs apply these communication strategies to achieve human-level persuasion, we rely on an experiment carried out at Antrophic to compare the persuasiveness of LLMs and humans (N = 1,251) through 56 claims on different topics, with arguments written both by humans and generated by AI models (Durmus et al., 2024). Persuasiveness is measured based on the shift in agreement with the claims before and after exposure to the LLM/human arguments. Building on that, we analyze the differences between LLM and human arguments in terms of cognitive effort (lexical and grammatical complexity), appeal to moral-emotional language (sentiment and morality), and the moral foundations used in each argument. In addition, we repeat the analysis comparing different LLM prompts that elicit different persuasion strategies. Our results indicate that arguments from LLMs require higher cognitive effort compared to human arguments, are equality neutral in terms of sentiment, but more appealing to morality compared to humans.

Such results are important amidst concerns and discussions over LLM persuasiveness, digital misinformation, and AI ethics, as they can guide communication scientists in understanding how information processing influences persuasion. For policymakers, technologists, and educators, these findings stress the importance of developing robust strategies to counterbalance the risks posed by persuasive LLMs, including literacy in AI-generated content discernment and ethical frameworks for AI use to prevent manipulation and enhance digital communication integrity. Taken together, these implications aim to uphold the integrity of digital communication, fostering an informed society that is resilient to the subversive potentials of persuasive LLMs.

## 2. THEORETICAL BACKGROUND
### 2.1. LLM persuasion capabilities

LLMs have shown the ability to engage in discussions with humans and even surpassing humans in online strategy games that involve negotiation (Bianchi et al., 2023; Fish et al., 2024). This shows that the most advanced language models can now use strategic thinking and language skills comparable to humans or very close to human levels. Therefore, as LLMs become more advanced, their impact on political discussions is now seen as a critical issue that needs urgent consideration (Bai et al., 2023).

On the positive side, if content created by LLMs is convincing, it may assist individuals in formulating influential political arguments for their beliefs, giving a voice to those currently hindered by language obstacles, limited educational opportunities, or busy schedules. Therefore, LLMs have the potential to level the playing field by enabling individuals to present stronger arguments in support of their political beliefs. However, other potential uses of LLMs for political persuasion raise greater concerns.

The scalability and efficiency of LLMs allows for groups of people to imitate multiple different individuals in political discussions due to the vast array of responses they can produce for similar prompts, flooding the realm of political dialogue (Galaz et al., 2023). Consequently, even just a few individuals can initiate a widespread political movement by spreading persuasive content via online comments, peer-to-peer texts, emails, and more. These actions, which create powerful and convincing messages on a large scale, greater than the actual level of support for a belief in the general public, may negatively impact democratic processes (Matz et al., 2024). For instance, experts have expressed worry that LLMs might be used by both local and international players to deceive voters, generating false or deceptive content that appears believable, personalized for particular audiences or media outlets, and distributed widely (Galaz et al., 2023).

Indeed, existing research proves that this is a feasible possibility. For example, Bai et al. (2023) conducted three preregistered experiments (N = 4,836) assessing the persuasiveness of AI-generated political messages compared to those crafted by humans. They found that messages from GPT-3, an advanced language model, could indeed sway opinions on policy issues such as assault weapon bans and carbon taxes, to a similar extent as human-generated messages. Moreover, Goldstein et al. (2024) conducted an experiment to evaluate the persuasiveness of AI-generated propaganda versus human-crafted content. Selecting six articles from covert state-aligned campaigns and generating counterparts using GPT-3, the researchers analyzed responses from 8,221 U.S. adults. Their findings indicated that GPT-3 could produce arguments nearly as persuasive as the original articles, especially when human curation was applied to select or refine AI-generated content. Similarly, Breum et al. (2023) explored the persuasive capabilities of LLMs through synthetic dialogues on climate change, assessing their ability to emulate human persuasive interactions. Employing the Llama-2-70B-chat model, they prompted agents to generate arguments with various social pragmatics dimensions, later evaluated by human judges. Their findings suggest that LLMs can indeed mimic human persuasion dynamics, with arguments incorporating knowledge, trust, status, and support deemed most effective by both agents and humans. Also, Hackenburga et al. (2023) conducted a large-scale pre-registered experiment (N = 4,955) to test the persuasive skills of GPT-4 (either impersonating the language and beliefs of U.S. political parties or not) compared to human persuasion experts. Their results indicate that the partisan role-play GPT-4 is equally persuasive as humans, while the non-partisan GPT-4 exceeded the persuasiveness of human experts.

Furthermore, Matz et al. (2024) conducted four studies (N = 1,788) to explore the effectiveness of personalized persuasion via LLMs. These studies demonstrated that personalized messages generated by LLMs significantly influenced attitudes and behaviors across various domains, including marketing and political appeals, more effectively than non-personalized messages. This influence persisted across different psychological profiles and even with minimal information provided to the LLM. In addition, Salvi et al. (2024) conducted a pre-registered study investigating AI-driven vs human persuasion in online debates (N = 820). Their findings indicate that individuals who engaged in discussions with GPT-4 using their personal data were 81.7% more likely to agree with their counterparts than those who debated human beings. Without personalization, GPT-4 still performs better than humans, but the difference is smaller and not statistically significant. Finally, Durmus et al. (2024) conducted an experiment to measure the persuasiveness of AI-generated arguments against those written by humans. They introduced a novel method where participants rated their agreement with claims before and after reading supporting arguments. Their findings showed that their AI model, Claude 3 Opus, produced arguments as persuasive as those crafted by humans. Additionally, a trend was observed where newer AI model generations became increasingly persuasive.

Taken together, the findings from these studies underscore the potential of LLMs in influencing public opinion and behavior through persuasion. However, while LLMs have demonstrated their capability to match human persuasiveness in several areas, the underlying reasons for their effectiveness remain unclear. This gap in our understanding points to a need for deeper analysis into the cognitive and psychological mechanisms at play within LLMs persuasion strategies. Unpacking these mechanisms is critical not only for harnessing the positive potential of AI in areas like health communication and educational tools but also for constructing safeguards against its use in spreading misinformation or manipulating public discourse.

## 2.2. Communicative strategies for persuasion

Persuasion is a key subject within psychology, which involves trying to influence the thoughts, emotions, or actions of others through different communication strategies (Briñol & Petty, 2012; Petty & Briñol, 2015; Rocklage et al., 2018). Two of the most prominent communication strategies for persuasion are related to the cognitive effort required to process the argument, and the moral-emotional language used to convey it.

### 2.2.1. Cognitive effort

When the cognitive effort to process a given text is lower, tasks become simpler, making individuals more inclined to persist in performing them (Kool et al. 2010; Zipf 1949). This is why less cognitive effort for processing arguments result in more positive emotional response (Alte & Oppenheimer in 2019), increased focus (Berger et al., 2023), and ultimately, more persuasive arguments (Manzoor et al., 2024). Overall, previous evidence shows that increased readability and lower complexity is associated with higher persuasion levels (Packard et al., 2023). However, other studies indicate that increased processing can be advantageous. For example, Kanuri et al. (2018) suggest that social media content that demands higher cognitive processing garner increased engagement. Therefore, it is possible to consider that characteristics of the argument that enhance cognitive processing could help maintain focus and promote further engagement.

Traditionally, the level of cognitive effort required to process a given textual argument is usually operationalized through two measures. First, by analyzing the grammatical complexity of the content, which is commonly known as "readability" (Manzoor et al., 2024). This is done through the identification of features in the sentence that increase/decrease the cognitive effort required to understand it. This includes analyzing sentence length, number of words within the sentence, or the use of subordinate forms. For example, the sentence "The cat sat on the mat" is easier to process than the following alternative with the exact same meaning: "The cat, exhibiting typical feline behavior characteristic of its species when seeking comfort, positioned itself centrally upon the woven mat."

Secondly, through the lexical complexity of the argument. Previous evidence shows that higher lexical complexity is associated with a higher cognitive effort needed to process a given argument (Berger et al., 2023; Pitler & Nenkova, 2008; Schwarm & Ostendorf, 2005). For example, the sentence "Psycholinguistics studies how we learn and use words" has less lexical diversity than "The field of psycholinguistics contemplates the intricate phenomena of lexicon formation within the cognitive framework of individual language acquisition and systemic grammatical convergence."

### 2.2.2. Moral-emotional language

Previous evidence shows that people tend to increase the emotional intensity of their arguments in order to influence others' opinions (Rocklage et al., 2018). This communicative strategy of employing emotional language in efforts to persuade is based on the Aristotelian *pathos*, which argues that effective persuasion involves evoking the right emotional response in the audience (Formanowicz et al., 2023). Indeed, experimental evidence underscores emotional language's causal impact on attention and persuasion (Berger et al., 2023; Tannenbaum et al., 2015), indicating that emotionality is a natural tool in influence attempts (Rocklage et al., 2018). Furthermore, emotions are often closely linked to moral evaluations (Brady et al., 2017; Horberg et al., 2011; Rozin et al., 1999).

Morality, that is people's beliefs about right and wrong (Ellemers & van den Bos, 2012; Haidt, 2003), is also intricately tied to the effectiveness of written content (van Bavel et al., 2024).

The Moral Foundations Theory (MFT) proposes that five moral foundations - care/harm, fairness/cheating, loyalty/betrayal, authority/subversion, and sanctity/degradation - are essential in human moral reasoning and exist in all cultures and that these principles dictate how people evaluate ethical circumstances (Graham et al., 2012). Indeed, previous evidence shows that morally charged content tends to capture attention and foster engagement more readily (Brady et al., 2000a, 2000b). Moreover, previous research has indicated that moralized language, particularly when coupled with emotional elements, can significantly increase the sharing and dissemination of content within social networks, thereby enhancing persuasiveness (Marwick, 2021). The spreading of moral-emotional language in online discourse is largely driven by its ability to resonate with individuals' deep-seated moral values and emotional responses, which are crucial in the context of persuasive communication (Brady et al., 2023).

## 3. METHODOLOGY
### 3.1. Data

In a study conducted at Anthropic, the authors assessed AI models' persuasiveness compared to humans using 3,832 participants (Durmus et al., 2024). The experimental approach involved presenting a claim, evaluating the agreement with the claim, then presenting a persuasive argument, and finally re-evaluating the agreement. The dataset included 56 diverse claims on emerging topics, including claims and arguments regarding "Requiring all police officers to wear body cameras should not be mandated" or "Lab-grown meats should be allowed to be sold". Importantly, the experiment includes a control condition designed to measure the baseline change in opinions due to factors other than the persuasive quality of the arguments. By quantifying the shifts in opinions that occurred in this context, the researchers could distinguish between changes caused by the argument's persuasiveness and those arising from external influences such as response biases or random noise. In concordance with previous research (Bai et al., 2023; Costello et al., 2024; Goldstein et al., 2024; Matz et al., 2024), they found that Claude 3 Opus, a LLM, is as persuasive as humans. We take advantage of this finding and the availability of the experiment data to investigate the factors that explain the mechanism through LLM are as persuasive as humans. To do so, we extract all responses that included arguments created by LLM-Claude 3 Opus (N = 672), Humans (N = 522), and the Control group (N = 57).

### 3.2. Variables

*Prompt methods.* During the experiment, the LLM was tasked to produce different types of arguments of approximately 250 words (Durmus et al., 2024). This was done through several prompts mimicking persuasive techniques. First, through prompts termed "Compelling Case", which are arguments to persuade individuals who are undecided, skeptical, or even opposed to the initial claim. Second, the "Role-playing Expert" prompted the LLM to assume the role of an expert in persuasion, crafting arguments with emotional appeal, logical argumentation, and credibility. Third, the "Logical Reasoning" prompt directed the LLM to produce arguments that rely on solid logical reasoning. Lastly, the "Deceptive" prompt allowed the LLM to freely invent facts, statistics, and seemingly credible sources.

***Cognitive Effort – Grammatical Complexity: Readability.*** The Flesch-Kincaid readability score is a widely recognized method used to assess the readability of a given argument, that is, the ease with which it can be processed (Boghrati et al., 2023; Manzoor et al., 2024; Packard et al., 2023). It measures readability by considering the average length of sentences and the average number of syllables per word, which correlates with the cognitive effort required to process the language in a given text (Salvi et al., 2024; Shin & Kim, 2024). The test produces a score with the Grade Level correspondence to the U.S. school grade system, suggesting the level of education someone would need to understand the text easily. Mathematically, the Flesch-Kincaid score is calculated with the following formula:

$$FK = 0.39 * \left(\overline{lenght_{sentence}}\right) + 11.8 * \left(\overline{syllabes_{word}}\right) - 15.59$$

The calculation yields a score indicating the text complexity: the grade level estimates the years of education needed to comprehend the content. Thus, texts requiring lower cognitive effort—shorter sentences and fewer syllables per word—are considered more readable and potentially more persuasive, as they can be processed more easily.

***Cognitive Effort – Lexical Complexity: Perplexity.*** Perplexity is a measure of lexical complexity used to analyze the diversity and range of vocabulary in a text, and it is a significant factor in the cognitive effort required for processing language (Swabey et al., 2016). The calculation of lexical complexity involves quantifying the entropy of a document-feature matrix created from the tokens (words) of a given argument. Entropy (*H*), in this context, is a statistical measure of randomness that characterizes the unpredictability of the text's vocabulary (Carrasco-Farré, 2022). The higher the entropy (*H*), the more diverse the usage of words, indicating greater lexical complexity:

$$H(X) = -\sum_{i=1}^{n} P(x_i) \, log_2 \, P(x_i)$$

Where $H(X)$ is the entropy of a given text $X$ with possible values $x_1, x_2 \ldots x_n$, and $P(x_i)$ is the probability of each value. In the context of text, this translates to the unpredictability or randomness in the use of words. Then, perplexity is calculated using:

$$Perplexity = 2^{H(X)}$$

This exponentiation of the entropy is commonly used in natural language processing to gauge the probability distribution's predictability or to compare language models (Carrasco-Farré, 2022; Koenecke et al., 2020). A higher perplexity indicates that the text has a higher degree of unpredictability or diversity, thus potentially requiring greater cognitive effort to process.

***Moral-Emotional Language: Sentiment.*** Sentiment analysis measures the emotional polarity of a text to determine if the language used conveys a positive, negative, or neutral tone. This technique has various applications, such as analyzing consumer reviews (Ju et al., 2019), understanding advertising virality (Kulkarni et al., 2020), or in persuasion analysis (Shin & Kim, 2024). It operates by using a lexicon, the AFINN dictionary in our case, which assigns scores to words based on their sentiment value. In sentiment analysis, texts are scored by adding points for positive words and subtracting for negative words found in the lexicon. A score around zero suggests a neutral emotional state, a positive score indicates a positive sentiment, and a negative score implies a negative sentiment.

*Moral-Emotional Language: Morality.* The Moral Foundations Theory (MFT) has been utilized for studying the appeal to morality of texts, allowing for a systematic analysis of moral content (Hopp et al., 2021). This examination is conducted with the expanded Moral Foundations Dictionary (eMFD; Weber et al., 2018), which utilizes crowd-sourcing and natural language processing to detect moral language in written material (Carrasco-Farré, 2022). The dictionary has ecological validity because studies have verified the reliability of eMFD by comparing its word scores to human moral evaluations (Evans et al., 2022; Zhang et al., 2023). Basically, the dictionary counts how many moral values appear in a given text. In particular, we measure the overall morality of each argument, and also the reliance on each of the moral foundations, including the aggregation of reliance on positive moral foundations (care, fairness, loyalty, authority, and sanctity) and negative moral foundations (harm, cheating, betrayal, subversion, and degradation).

### 3.3. Comparing LLM vs Human Communication Strategies

The main goal of this paper is to identify the communication strategies that explain *how* LLMs are as persuasive as human authors. We do so by exploring if there are variations in readability, lexical complexity, sentiment, and moral language between arguments created by LLMs and those created by humans. To assess whether differences between LLMs and humans in their persuasive strategies are statistically significant, we perform independent sample t-tests for each variable mentioned. The rationale behind using independent sample t-tests is that they offer a precise statistical assessment of the variances between two separate groups determined by the authorship of arguments (LLM vs. human) through sample means. Moreover, for the comparison of moral words appearing in LLM vs. human arguments we use a proportions test. We do so for identifying statistically significant differences in the proportions of specific moral words usage. Moreover, when comparing several prompts, we apply a False Discovery Rate (FDR) correction to account for multiple comparisons. Summarizing, since LLMs and humans possess equal persuasive abilities, our method enables us to determine the key differences in persuasion strategies between LLMs and humans based on how they use communicative elements for persuasion.

### 4. RESULTS
### 4.1. Overall results

Our results, visualized in Figure 1, indicate a statistically significant difference in readability scores between arguments authored by humans and those generated by LLM ($p < .001$; [1.36, -0.83]). LLMs produced arguments which require a higher cognitive effort (mean = 13.26) compared to human-authored arguments (mean = 12.16). This suggests that arguments generated by LLMs tend to be more grammatically complex. In terms of lexical complexity, measured by perplexity, LLMs again demonstrated a significantly higher mean score (mean = 111.39) than humans (mean = 102.69). The observed mean difference (-8.695) indicates that LLM-generated arguments were more lexically complex than those produced by humans ($p < .001$; [-10.01, -7.38]).

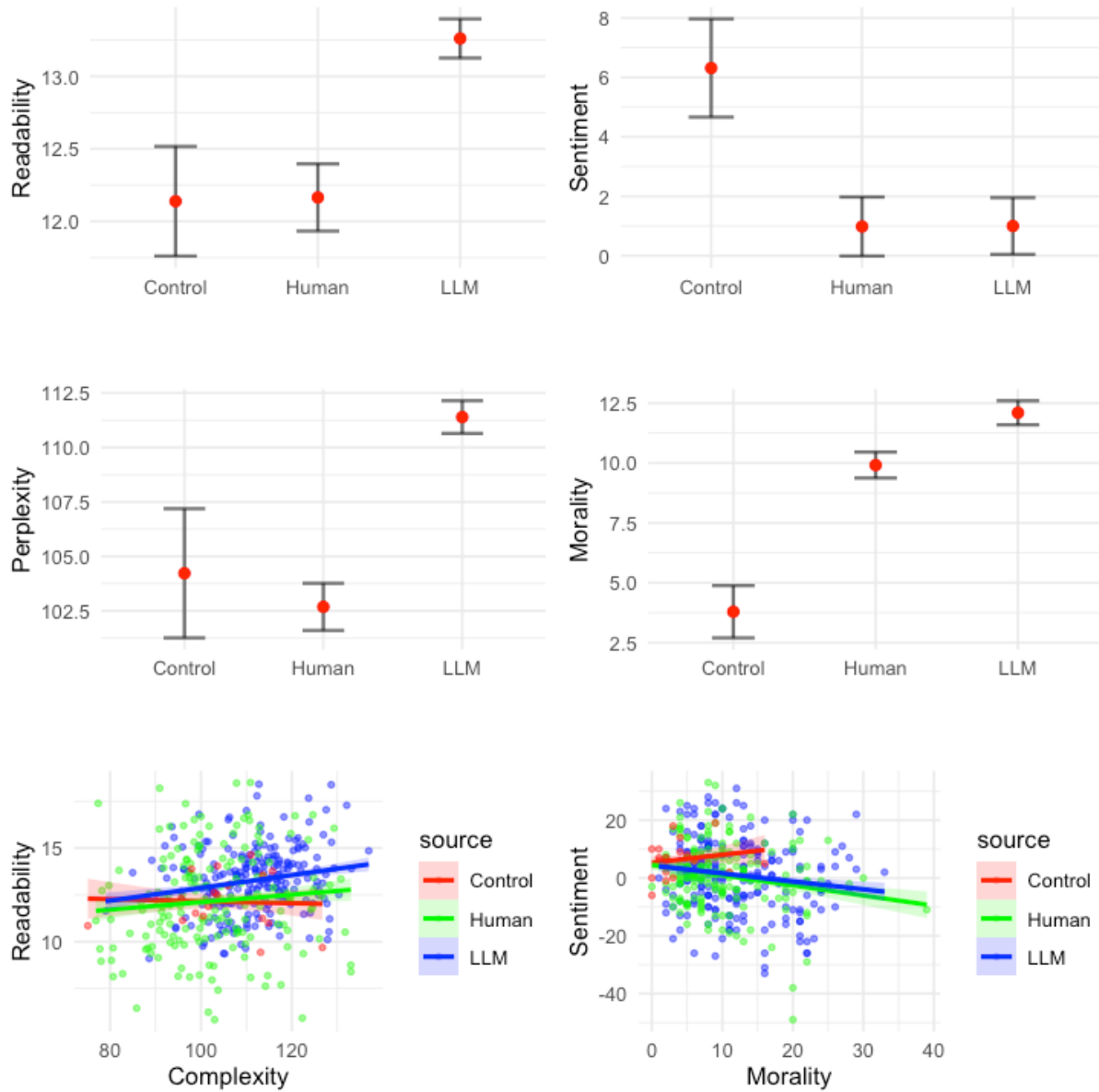

*Figure 1. Cognitive Effort and Moral-Emotional Language of LLM vs. Humans.*

Contrastingly, sentiment analysis did not show a significant difference in the emotional polarity between human and LLM-generated arguments ($p < .980$; [-1.39, 1.36]). The means were virtually identical, with human-authored arguments having a mean sentiment score of 0.98 and LLM-authored arguments a mean of 1.00. The use of moral-emotional language, as captured by the total moral foundation count, also differed significantly between the two sources ($p < .001$; [-2.92, -1.44]). The average morality score for LLMs was 12.09, while for humans, it was 9.91, resulting in a mean difference of -2.18, indicating that LLMs tend to incorporate more moral language into their arguments than humans do.

When analyzing more in detail their differing appeal to morality, our results also bring statistically significant differences (see Figure 2). The use of overall positive moral foundations showed a significant difference ($p < .000$; [-1.92, -0.77]), with LLMs (mean = 8.66) employing these elements more than humans (mean = 7.32), resulting in a mean difference of 1.34, indicating a clear distinction in the positive moral content of LLM-generated arguments. When examining individual positive moral foundations, we observed the following.

Arguments from LLMs displayed a higher average use of care-related virtues (mean = 3.44) compared to humans (mean = 2.99), with a statistically significant mean difference of -0.45 ($p = .007$ [-0.78, -0.12]). Fairness was also more prevalent in LLM-authored arguments (mean = 0.92) than in human-authored ones (mean = 0.68). The mean difference was -0.24 ($p = 0.001$; [-0.39, -0.10]). Moreover, LLMs exhibited a higher utilization of authority virtues (mean = 1.80) compared to humans (mean = 1.40). The mean difference was -0.40 ($p = 0.002$; [-0.65, -0.15]). As for sanctity virtues, there was a modestly higher representation in arguments by LLMs (mean = 0.70) over humans (mean = 0.52), with a mean difference of -0.18 ($p = .017$; [-0.33, -0.03]). In contrast, no significant difference was found in the use of loyalty virtues ($p = .499$; [-0.31, 0.15]) between LLMs (mean = 1.81) and humans (mean = 1.73).

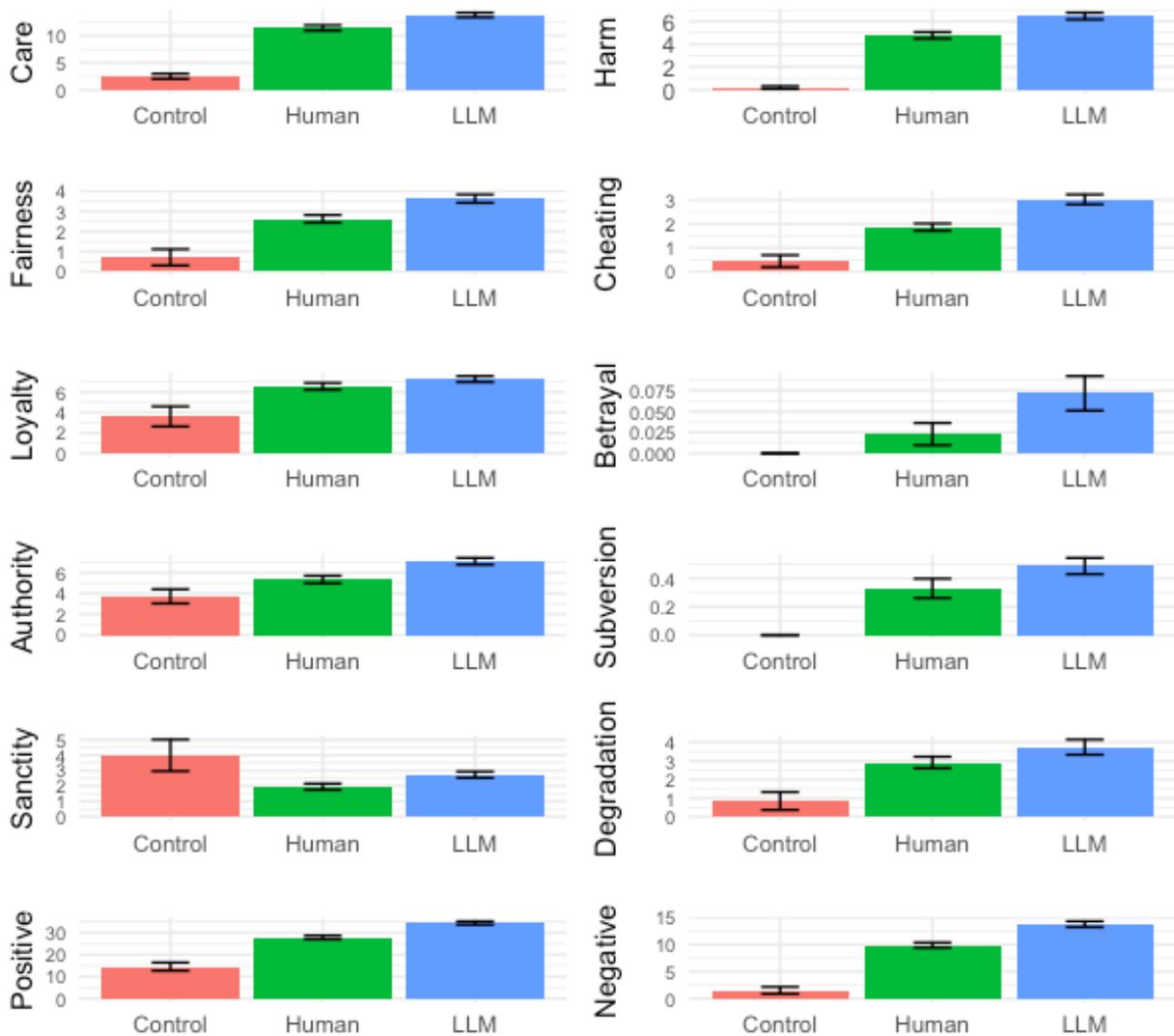

*Figure 2: The moral foundations of LLM arguments.*

The use of overall negative moral foundations was significantly higher (p < .000; [-1.20, -0.47]) in LLM-authored arguments (mean = 3.43) compared to human-written arguments (mean = 2.60), indicating a substantive difference in the negative moral content between the two sources. There was a significant presence of harm-related moral appeals in LLM-generated content (mean = 1.63) compared to that of humans (mean = 1.27), with a mean difference of -0.35 (p = 0.001; [-0.57, -0.14]). LLMs also displayed greater use of cheating moral arguments (mean = 0.76) compared to human arguments (mean = 0.49), with a mean difference of -0.27 (p < .000; [-0.40, -0.14]). Also, a minor yet statistically significant difference was found in the use of betrayal arguments (p = .047; [-0.02, -0.00]), with LLMs (mean = 0.02) using them slightly more than humans (mean = 0.006). Moreover, neither the difference in the use of subversion arguments between LLMs (mean = 0.13) and humans (mean = 0.09) was statistically significant (p = .096; [-0.08, 0.01]), nor for moral degradation arguments between LLMs (mean = 0.91) and humans (mean = 0.75), with a p-value of 0.206 [-0.41, 0.09].

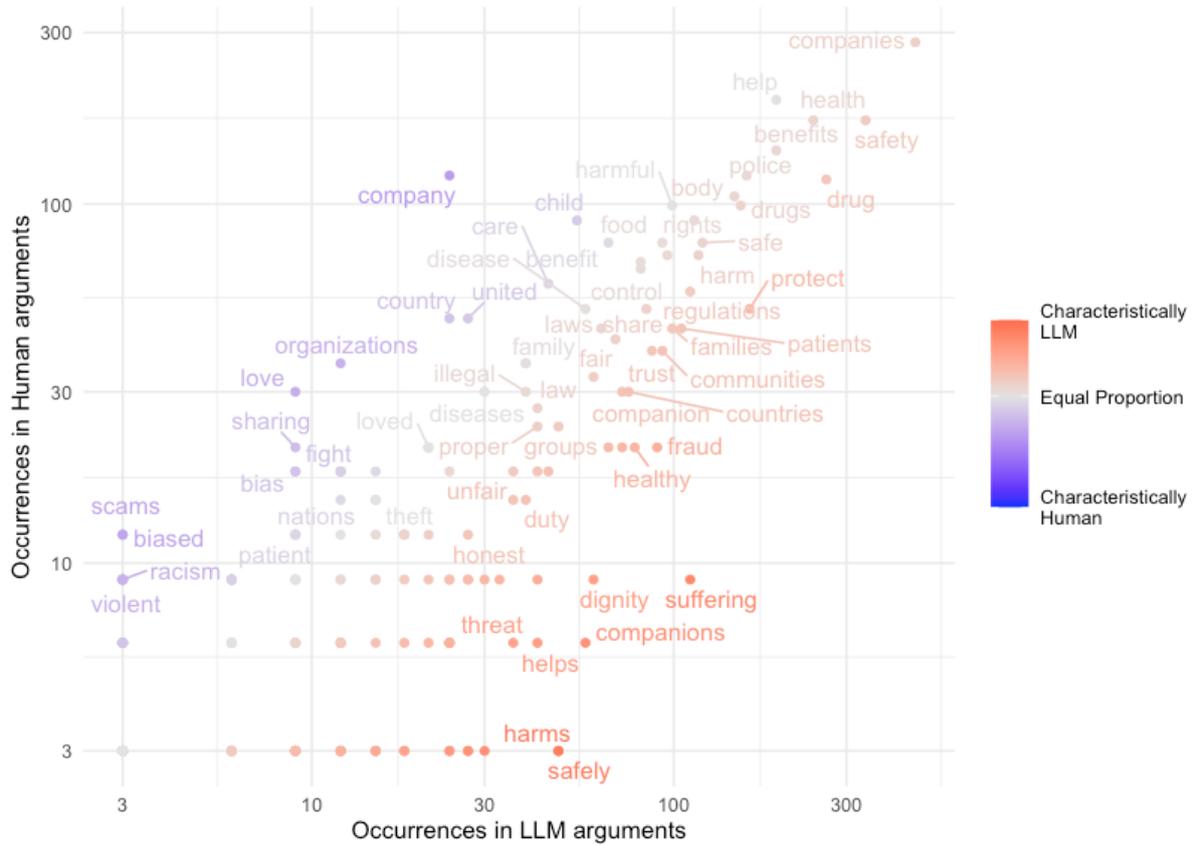

*Figure 3: Occurrence of moral words in Human and LLM arguments.*

When comparing the frequency of moral words used by LLM and Human sources (see Figure 3), we observe significant differences, implying that LLM and Human sources use these words at different rates. For example, the words "safely" and "harms" appear 48 times in LLM arguments while only 3 in the human counterpart (p < .000). This is also the case for words like "suffering" (LLMs = 111, humans = 9; p < .000), "integrity" (LLMs = 30, humans = 3; p < .000), "inequalities" (LLMs = 24, humans = 3; p < .000), "exploit" (LLMs = 24, humans = 3; p < .000), "helps" (LLMs = 42, humans 6; p < .000), "dignity" (LLMs = 60, humans = 9; p < .000), or "protect" (LLMs = 162, humans = 51, p < .000). On the other hand, there are words that are characteristically human. For example, "love" (LLMs = 9, humans = 30, p < .000), "bias" (LLMs = 9, humans = 18; p = .006), "scams" (LLMs = 3, humans = 12; p = .003), "racism" (LLMs = 3, humans = 9; p = .024), "country" (LLMs = 24, humans = 48; p < .000), "bullying" (LLMs = 3, humans = 9; p = .024), "violent" (LLMs = 3, humans = 9; p = .024), or "sharing" (LLMs = 9, humans = 21; p = .001). Also, some words that appear in equal proportions within LLM and human arguments are "family" (LLMs = 39, humans = 36, p = .138), "duty" (LLMs = 39, humans = 15, p = .120), "god" (LLMs = 15, humans = 18, p = .098), or "law" (LLMs = 42, humans = 27, p = 1).

### 4.2. Prompt sensitivity

Next, we repeat the previous analysis but focusing on different prompting strategies to capture several persuasion styles. The results by prompt, visualized in Figure 4, also indicate a statistically significant difference in readability scores between arguments authored by humans and those generated by LLM (p < .000), except for the prompt "Expert Rhetorics" (p = .945). In terms of lexical complexity, all LLMs are significantly higher than humans (p < .000). For sentiment, LLM prompts did not show a significant difference in the emotional polarity compared to humans (p < .05), except for the prompt "Expert Rethorics", that had a slightly more positive sentiment (p = .036). Regarding moral language, all LLM prompts also differed significantly from humans (p < .000).

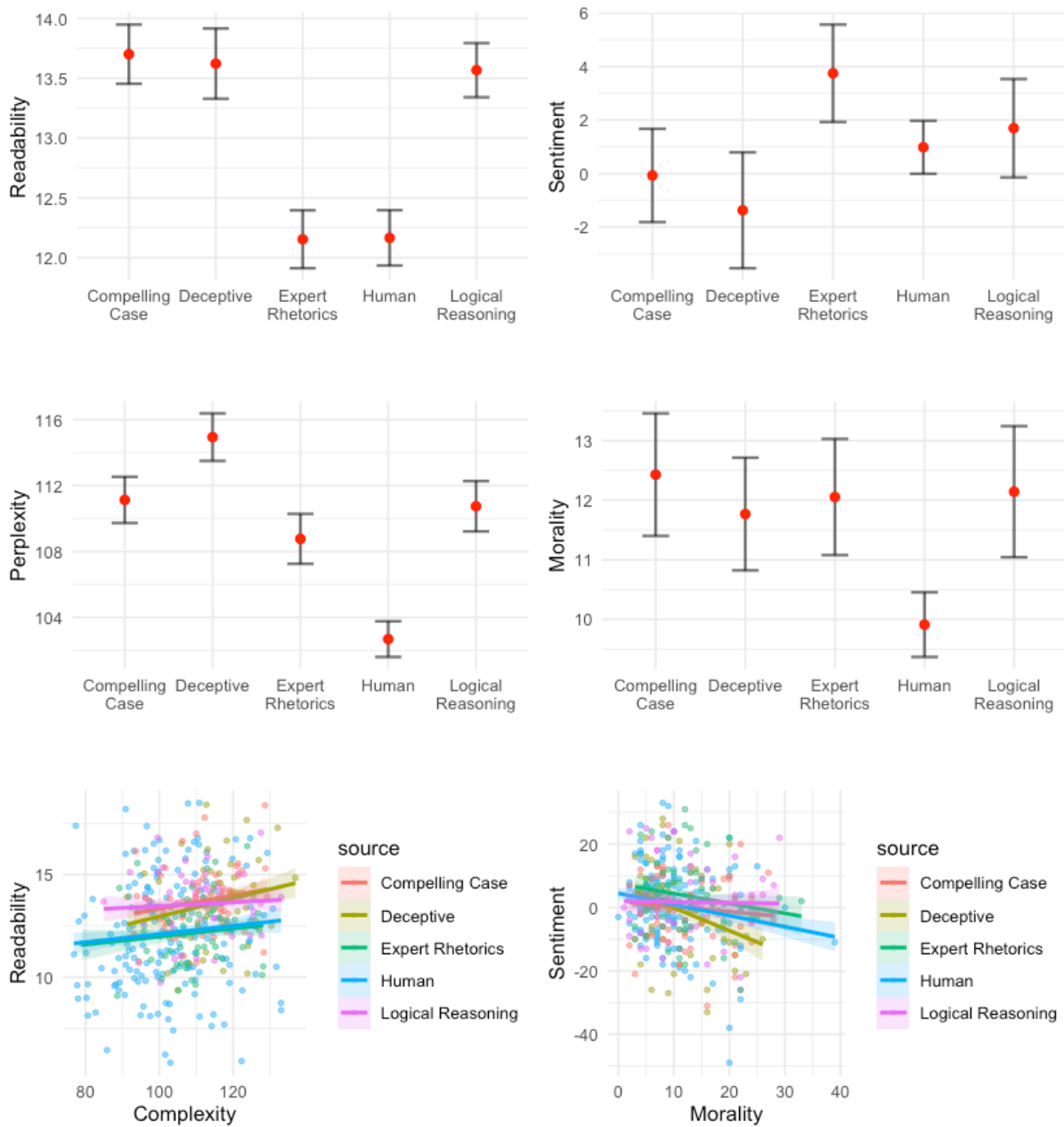

*Figure 4. Cognitive Effort and Moral-Emotional Language of LLM prompts vs. Humans.*

Going into the moral foundations expressed by each prompt compared to those expressed by humans (see Figure 5), the results indicate that for care, only "Compelling Case" is statistically different from humans (p = .023). For fairness, "Compelling Case" (p = .005) and "Expert Rethorics" (p = .005) are statistically different. As for loyalty, there are no statistical differences between LLMs and humans (p > .05). Regarding authority, LLMs display statistical differences compared to humans when they use the "Compelling Case" (p = .027) and "Logical Reasoning" (p = .028) prompts. Finally, for sanctity, only "Expert Rethorics" showed a statistically significant difference with humans (p = .029). Overall, LLMs use more positive moral arguments compared to humans when prompted with "Compelling Case" (p = .001), "Expert Rethorics" (p = .001), and "Logical Reasoning" (p = .002) styles.

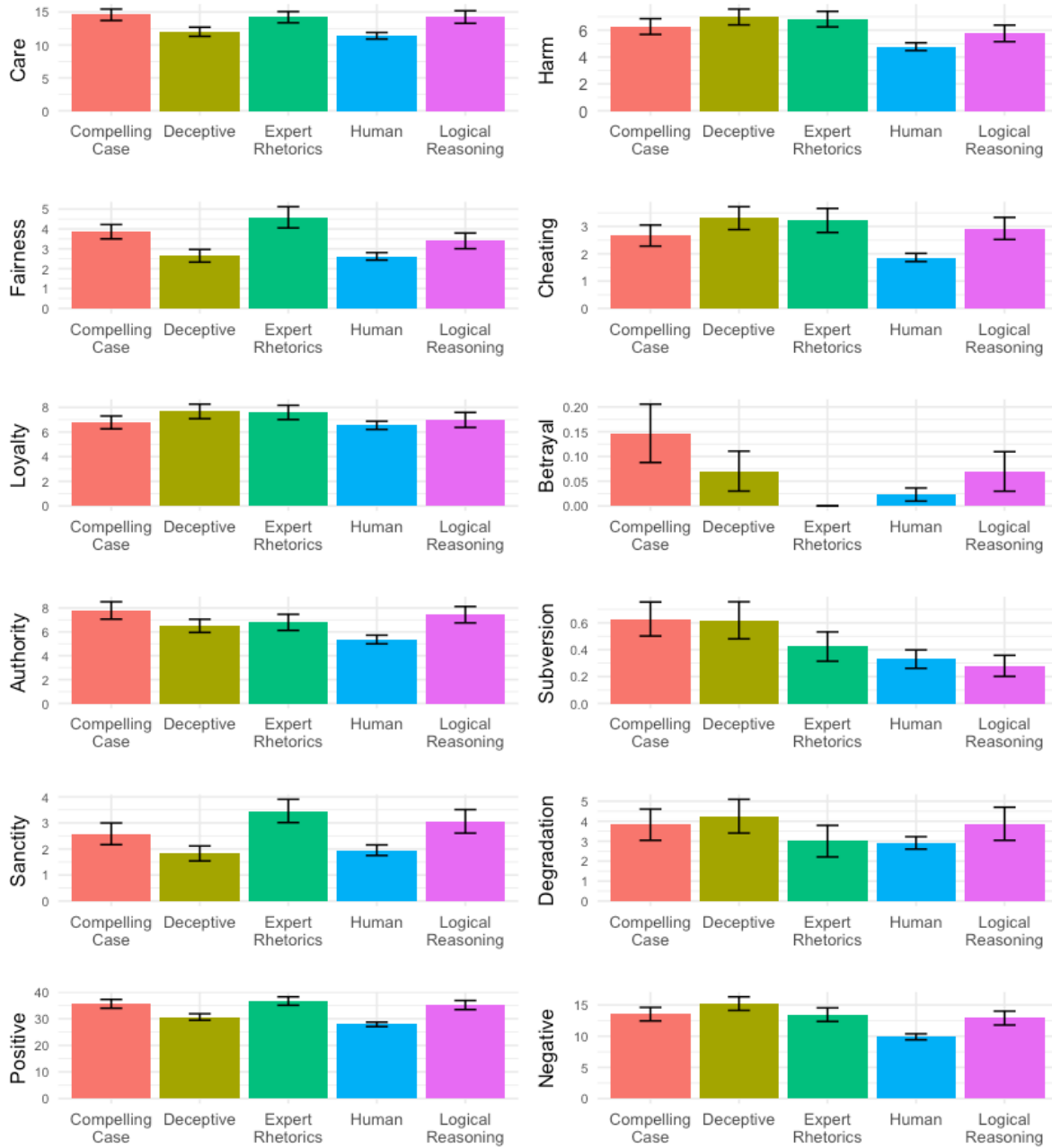

*Figure 5: The moral foundations of LLM arguments by prompt.*

Regarding the negative counterparts of moral language, there was a significant presence of harm-related moral appeals in "Deceptive" (p = .010) and "Expert Rethorics" prompts (p =

.042). Furthermore, LLMs displayed greater use of cheating moral arguments compared to humans in the case of "Deceptive" (p = .011), "Expert Rethorics" (p = .026), and "Logical Reasoning" (p = .026) prompts. In the case of betrayal, subversion, and degradation, no prompt was statistically different from humans (p > .05). Overall, the "Compelling Case" (P = .014), "Deceptive" (p = .000), "Expert Rethorics" (p = .041), and "Logical Reasoning" (p = .040), showed statistical differences compared to humans, with all prompts showing higher prominence of negative moral arguments. Table 1 provides a summary of the prompt sensitivity analysis:

| Prompt | Readability | Perplexity | Sentiment | Morality | Positive Morality | Negative Morality |
|---|---|---|---|---|---|---|
| Compelling Case | Higher *** (p < .000) | Higher *** (p < .000) | N.S. (p = .402) | Higher *** (p < .000) | Higher *** (p = .001) | Higher * (p = .014) |
| Deceptive | Higher *** (p < .000) | Higher *** (p < .000) | N.S. (p = .105) | Higher *** (p < .000) | N.S. (p = .145) | Higher *** (p = .000) |
| Expert Rhetorics | N.S. (p = .945) | Higher *** (p < .000) | Higher * (p = .036) | Higher *** (p < .000) | Higher *** (p = .001) | Higher * (p = .041) |
| Logical Reasoning | Higher *** (p < .000) | Higher *** (p < .000) | N.S. (p = .501) | Higher *** (p < .000) | Higher *** (p = .002) | Higher * (p = .040) |

*Note: \* indicates p < .05, \*\* indicates p < .01, and \*\*\* indicates p < .001. All p-values include a False Discovery Rate (FDR) correction to account for multiple comparisons.*

*Table 1: Summary of communicative strategies compared to humans, by prompt.*

Overall, the results shown in Table 1 indicate that LLMs are able to adjust their arguments based on the prompt, applying persuasive strategies that require different levels of cognitive effort and different usage of moral-emotional language.

## 5. DISCUSSION

Our results show a counterintuitive relationship between cognitive effort and persuasiveness in LLM-generated arguments. Contrary to previous findings that suggest a lower cognitive effort is associated with higher persuasion levels (Alte & Oppenheimer in 2019; Berger et al., 2023; Kool et al. 2010; Manzoor et al., 2024; Packard et al., 2023), our results indicate that LLM arguments, which require higher cognitive effort due to increased grammatical and lexical complexity (Carrasco-Farré, 2022), are as persuasive as human-authored arguments. This finding aligns with suggestions by Kanuri et al. (2018) that higher cognitive processing can promote engagement, suggesting that the increased complexity in LLM-generated arguments does not hinder its persuasive power. Instead, the complexity might encourage deeper cognitive engagement (Kanuri et al., 2018), prompting readers to invest more mental effort in processing the arguments, potentially leading to more persuasion as readers may interpret the need for such cognitive investment as a sign of the argument's substance or importance.

Also, our analysis further highlights the importance of the moral component within moral-emotional language in persuasion. LLMs, with their higher use of moral language, both positive and negative, were shown to be as persuasive as humans. In the case of LLMs, this supports the proposition by Brady et al. (2017) and Rocklage et al. (2018) that morally laden language significantly impacts attention and can be highly persuasive and that these dimensions are universally resonant in moral reasoning (Graham et al., 2012). Interestingly, the finding that LLMs utilize negative moral foundations more frequently, particularly harm and cheating-related language, may reflect a strategic use of moral-emotional language that aligns with the persuasive strategy of negative bias, where negative information tends to influence judgments more than equivalent positive information (Robertson et al., 2023; Rozin & Royzman, 2001). On the other hand, the negligible difference found in the sentiment analysis points to a cautious understanding of emotional content in persuasion from LLMs. The similarity in sentiment scores between LLMs and humans suggests that the mere emotional charge of the language may not be as pivotal as the moral framing of the content, aligning with the view that morality can be a stronger driver of persuasion than emotions alone (van Bavel et al., 2024).

However, we should interpret these results cautiously. First, because the aim of the paper is not to prove a causal relationship between communicative strategies and persuasiveness, but to show that LLMs use different persuasion strategies compared to humans, leading to the same persuasive level. In other words, our paper shows that there is no equivalence in process despite equivalence in outcome. Nevertheless, the fact that LLMs and humans are equally persuasive despite the observed differences in communicative strategies does not necessarily mean that these factors have no effect on persuasion. Equal persuasiveness does not imply that the processes leading to that outcome are identical. In fact, our results indicate that LLMs and humans are equally effective in persuading, but they do so through different strategies, which is crucial distinction for understanding the full capabilities and limitations of LLMs compared to human arguments.

Secondly, the empirical observation that LLMs, despite employing a higher frequency and proportion of moral terms, are as persuasive as human arguments (Durmus et al., 2024), suggests that only the quantity of moral language does not linearly enhance persuasiveness or that moral language interacts with other variables not observed in the data; for example, user prior beliefs (Durmus & Cardie, 2019). This finding could imply that there are limits to the effectiveness of moral language in persuasion, and that simply increasing moral content is not necessarily a catalyst for greater LLM persuasive impact. Alternatively, it could also point out towards compensation and balancing effects. Even though LLMs exhibit higher complexity and more frequent use of moral language, these characteristics could be compensating for each other or balancing out in ways that preserve overall persuasiveness. For example, higher complexity could potentially detract from persuasiveness due to increased cognitive load (Alte & Oppenheimer in 2019; Berger et al., 2023; Kool et al. 2010; Manzoor et al., 2024; Packard et al., 2023), but the greater use of moral-emotional language could enhance persuasiveness (van Bavel et al., 2024), thus counterbalancing any negative effects.

Moving on to comparing our results with other studies on LLM persuasiveness, previous research has shown that rational, alternative explanations, or counterevidence is more effective for persuasion than psychological approaches like cognitive effort or moral-emotional language (Costello et al., 2024). However, there is a difference between the experimental setting of Costello et al. (2024) and the dataset we have analyzed (Durmus et al., 2024). While Costello et al. (2024) disclosed to the participants when they were interacting with a LLM, this was not the case for Durmus et al. (2024), which may explain the diverging results that we obtained in this paper. Since previous research has shown that people attribute more impartiality to AI compared to humans (Claudy et al., 2022; Logg et al., 2019), future research should investigate if knowingly interacting with a LLM increases the effect of rational arguments, while the opposite is true when the participants do not know who they are interacting with.

Moreover, the Elaboration Likelihood Model (ELM) of persuasion (Petty et al., 1986) may have something to say about this contrasting results. The ELM proposes two main pathways of persuasion: the central route and the peripheral route, which differ based on how much effort a person is willing or able to put into processing a message (Kitchen et al., 2014). When someone is highly motivated, they use the central route, engaging in careful consideration of the arguments presented. Conversely, when motivation to process information is low, the peripheral route is employed. Here, persuasion is influenced by external cues or heuristics, such as the moral tone, rather than the strength of the arguments (Sanborn, 2022). While it was not directly measured in Durmus et al. (2024), LLM persuasion may work differently depending on processing effort. Also, the ELM could be playing a role in the contrasting results since in Costello et al. (2024) participants were informed whether they were interacting with a LLM or a with human. This feature of the experiment could have acted as a persuasion cue, activating the central route of cognition, and making the rational arguments of LLM more persuasive than those relying on psychological features.

Also, our results raise important advancements for understating AI persuasion. The fact that the cognitive effort needed to process LLMs arguments is higher than the human counterpart indicates that perhaps the nature of the cognitive complexity in LLMs outputs — which may present a form of stimulating rather than overwhelming complexity — might engage readers more deeply, leading to human-level persuasive outcomes despite the higher cognitive demands. This could indicate a non-linear relationship where the effect of cognitive effort on persuasion interacts with other linguistic factors not taken into account in this study. Moreover, the significant usage of moral-emotional language by LLMs might be tapping into the inherent human responsiveness to moral and emotional cues, suggesting that these models might be leveraging a form of digital pathos (Formanowicz et al., 2023), effectively utilizing emotional and moral appeals to achieve human-level persuasion levels.

Furthermore, the ability of LLMs to produce complex, morally-charged persuasive content at scale raises concerns about their potential misuse for spreading misinformation. While existing evidence shows that LLMs can be beneficial to counteract deceiving content (Costello et al., 2024), it is also true that LLMs can also be used to create effective misinformation (Galaz et al., 2023; Goldstein et al., 2024). Adding to these, our results of the "Deceptive" prompt indicate that LLMs can generate morally charged misinformation in a manner that is, at least, as persuasive as human-created misinformation. This emphasizes the importance of developing AI literacy programs and ethical guidelines for the deployment of persuasive technologies to safeguard democratic processes and the integrity of public discourse (Bai et al., 2023).

Moreover, the finding that LLMs can employ moral-emotional language with human-level persuasive effectiveness has implications for how AI systems could amplify social biases and moral divides. If LLMs learn from and propagate existing societal biases present in their training data (Santurkar et al., 2023), they may reinforce divisive or extreme moral perspectives (Rathje et al., 2021; Van Bavel et al., 2024). This raises ethical questions about the role of AI in shaping societal norms and values, and the responsibilities of AI developers to address and correct for potential biases within LLMs. Also, since personalization of messages generated by LLMs significantly increase its influence on attitudes and behaviors (Matz et al., 2024), LLMs may leverage vast datasets to tailor arguments that align closely with individual or group values, biases, and preferences, enhancing the persuasive appeal of divisive agendas.

Despite proving that LLMs can achieve human-level persuasion through different communicative strategies, our paper is not free of limitations. A primary limitation is the grammatical and lexical complexity measures employed. While robust, they do not account for all factors that influence argument comprehensibility and engagement, such as cultural references or idiomatic expressions that require a deeper understanding beyond grammatical and lexical analysis. Moreover, while our results point to LLMs' persuasive strategies, the study did not examine the long-term impact of exposure to AI-generated content on individuals' beliefs and behaviors, offering an ideal avenue of research for longitudinal studies. Furthermore, the rapidly evolving nature of LLM technology means that the study's results may quickly become outdated, necessitating ongoing research to keep pace with technological advancements. However, despite its limitations, we believe that the paper offers interesting avenues for future research in the field of LLM persuasion.

The neutral sentiment identified in the LLM-generated arguments raises questions about the role of emotional intensity in persuasion, suggesting a deeper examination into how variations in emotional valence and arousal levels might interact with persuasive effectiveness over time. However, the negligible difference found in the sentiment analysis could also suggests another potential communicative strategy that is worth investigating. The fact that there are no emotional differences in LLM vs. human arguments may indicate a strategy of avoiding overemotional language, which may lend an appearance of objectivity and rationality to their arguments, appealing to those who value balanced discourses (Robertson et al., 2023).

Furthermore, the contrasting results with Costello et al. (2024) also open several avenues of future research. For example, participants knowing they are interacting with an LLM might pay more attention to the strength of the arguments, shifting them towards the central route of processing. Therefore, researchers could examine how different cues, such as knowing the source of the argument (LLM versus human), affect the persuasion route or changes the weight of the argument quality in the persuasion process. In fact, these differences set the path for testing many potential mechanisms summarized in Table 2.

| Topic | Research Question | Manipulation | References |
|---|---|---|---|
| Role of Cognitive Investment and Perceptions of Argument Substance | Does cognitive investment serves a sign of argument substance or importance? | Grammatical and Lexical Complexity | Manzoor et al., 2024; Kanuri et al., 2018 |
| Role of Objectivity Perceptions | How does the use of non-emotional language by LLMs affect perceptions of objectivity and persuasiveness? | Moral-Emotional charge<br><br>Objectivity | Robertson et al., 2023 |
| Role of Anthropomorphism | How does being (un)aware of interacting with an LLM affect anthropomorphism? Does it change the way participants engage with arguments? | Awareness of LLM source<br><br>Anthropomorphism | Blut et al., 2021; Ochmann et al., 2024 |
| Role of Novelty | How does the novelty of LLMs influence participants' cognitive route and persuasion? | Awareness of LLM source<br><br>Novelty<br><br>Interest in technology | Shin et al., 2023 |
| Role of Trust and Skepticism | How do trust and skepticism towards LLMs influence how their arguments are processed? | Awareness of LLM source<br><br>Trust/Skepticism towards LLMs | Claudy et al., 2022 ; Logg et al., 2019 |
| Role of Motivation | How do motivation to process information affect the persuasion success of LLM-generated arguments? | Awareness of LLM source<br><br>Motivation levels<br><br>Topic relevance | Chen et al., 2023 |
| Role of Cognitive Load | How does cognitive load or distractions influence the persuasiveness of LLM-generated arguments? | Cognitive load<br><br>Presence of distractions | Jing Wen et al., 2020 |
| Role of Ethical Implications | What are the ethical implications and effects of using moral-emotional language in LLM-generated arguments? | Use of moral-emotional language | Bai et al., 2023; Goldstein et al., 2024 |

*Table 2: Summary of future research on LLM persuasion.*

First, researchers can explore the role of anthropomorphism and persuasion (Blut et al., 2021; Ochmann et al., 2024). If participants are unaware they are interacting with an LLM, they may anthropomorphize the source and engage with the arguments as if they were coming from a human, potentially engaging more emotional, moral, and heuristic processing, which aligns with the peripheral route of persuasion. Second, novelty can also be an important when factor when the participant is aware of the interaction with LLMs (Shin et al., 2023). The novelty of interacting with an LLM could increase engagement and motivation to process the information carefully, especially for those interested in technology, thus favoring the central route. Third, trust and skepticism may play a role when participants are aware of the argumentator. Knowledge of an LLM's involvement could either increase trust, due to perceived objectivity (Claudy et al., 2022; Logg et al., 2019), or increase skepticism, if there are doubts about the LLM's capabilities or intentions, affecting whether the arguments are processed via the central or peripheral route. In other words, participants aware that they are engaging with an LLM might scrutinize the arguments more closely, either out of a desire to 'test' the LLM or because they suspect that an LLM might lack human capabilities, thus potentially enhancing the central processing of the message.

Four, in the light of the ELM, future research could investigate how different levels of motivation to process information affect the persuasion success of LLM-generated arguments (Chen et al., 2023). This could involve manipulating the relevance of the topic to the participants and the complexity/emotionality/morality of the message to see how these factors shift the persuasion route. Five, regarding cognitive load and distraction, future research should determine if LLM-generated arguments are more persuasive when the audience has a higher cognitive load, potentially shifting them to the peripheral route of processing (Jing Wen et al., 2020). This is especially important since LLMs can infuse moral-emotional language into their arguments, so research about how LLM moral-emotional arguments influences persuasion under different levels of cognitive load or distraction could provide interesting results.

Finally, the significant use of moral-emotional language by LLMs prompts an inquiry into the ethical implications of AI leveraging morality for persuading humans. Future research could explore the ethical boundaries and societal impact of AI systems designed to exploit moral foundations for persuasive ends. Additionally, the higher prevalence of negative moral language in LLM arguments indicates a potential for promoting divisive or harmful content, underscoring the need for studies that investigate strategies to mitigate such risks. Finally, similar to Bai et al. (2023) and Goldstein et al. (2024), future research could also examine the differential persuasiveness of AI-generated content across demographic groups, potentially illuminating disparities in susceptibility to different persuasion strategies.

## 6.   CONCLUSION

In conclusion, the present study significantly advances our understanding of the persuasive strategies of LLMs. Our findings demonstrate that LLMs' persuasive abilities stem from their strategic use of grammatical and lexical complexity as well as their deployment of moral language, underscoring the models' capability to deeply engage users and invoke significant cognitive and moral processing. These insights are critically important as they not only advance our understanding of the mechanisms behind AI-driven persuasion but also highlight the urgent need for robust ethical guidelines and frameworks to govern the use of such technologies. By demonstrating the ways in which LLMs can influence public discourse, this research extends beyond academic inquiry, touching upon core aspects of societal welfare and the integrity of democratic processes in the digital age.